# KurdSTS: The Kurdish Semantic Textual Similarity

Abdulhady Abas[1,2], Hadi Veisi[3], Hussein Mohammed Ali[1]

[1]Artificial Intelligence and Innovation Centre, University of Kurdistan Hewler, Erbil, Iraq. (abdulhady.abas@ukh.edu.krd, hussein.mohammedali@ukh.edu.krd)

[2]Computer Science Department, Faculty of Science, Soran University, Soran, Irbil, Kurdistan, Iraq.

[3]Faculty of New Sciences and Technologies, University of Tehran, Tehran, Iran. (h.veisi@ut.ac.ir)

## Abstract

Semantic Textual Similarity measures the degree of equivalence between the two texts and is important in many Natural Language Processing tasks. While extensive resources have been developed for high-resource languages, unfortunately, low-resource languages, for example, Kurdish, have been neglected. In this paper, the first STS dataset for Kurdish has been introduced, which aims to alleviate this gap. This dataset contains 10,000 formal and informal sentence pairs annotated for similarity. To this end, after benchmarking several models, such as Sentence Bidirectional Encoder Representations from Transformers (Sentence-BERT) and multilingual Bidirectional Encoder Representations from Transformers (multilingual BERT), among others, which achieved promising results while also showcasing the difficulties presented by the distinctive nature of Kurdish. This work paves the way for future studies in Kurdish semantic research and Natural Language Processing in general for other low-resource languages.

## Data Online: Link.

## 1. Introduction

Natural Language Processing (NLP) represents an artificial intelligence application that enables computers to process and create human language. The most challenging NLP application is Semantic Textual Similarity of measuring semantic text equivalence (Reimers, 2019). The wide range of possible applications for Natural Language Processing includes question answering and information retrieval but it provides its most meaningful transformation in specialized yet essential fields. Plagiarism stands as an essential domain for academic as well as professional integrity (El-Muwalla and Badran, 2020). The abundance of plagiarism detection tools exists only in high-resource languages such as English but the Kurdish language especially its Central Kurdish dialect (CKB) receives no attention (Veisi et al., 2024). The lack of computational support for Kurdish language threatens both intellectual honesty maintenance and further development of low-resource language processing.

Kurdish language is believed to be one of the oldest languages in Indo-European family; It is spoken by millions of people in the Kurdistan region which is a part of Iraq and Iran, Türkiye, and Syria (Khalid, 2020). Central Kurdish (Sorani) is a valuable highly inflected morpheme, phrase, and sentence structural of the given language and is distinctively preserved in Arabic letters. At the same time, its richness makes it difficult for computational analysis (Abdullah et al., 2024). Central Kurdish is the first language of millions of people; however, there is no dataset or tool available for STS, let alone for plagiarism detection. Consequently, academic professionals, educators, and publishers within Kurdish-speaking communities are obliged to depend on manual, time-consuming, and inconsistent methods that are out of sync with the digital era.

Moreover, the inability to detect semantic repetition and paraphrase discriminatorily hinders the advancement of scholarly work and the development of new technological tools.

What gaps does this paper aim to address? This paper takes the first steps in providing foundational work as the first in the field to advance development. One of the goals involves the collection of the first-ever Central Kurdish STS data. This is primarily to facilitate effective STS in the Kurdish language. By concentrating on this critical application, we aim to establish a foundation that will, in the most positive way, address the demands of Central Kurdish Natural Language Processing (NLP). This data captures the language intricacies of Central Kurdish, particularly in its various intricate syntactic arrangements as well as its different morphological structures. This ensures that Central Kurdish NLP speech technology, currently dominating with non-functional tools, will not only be functional but also linguistically smart. It provides scaffolding to a system that will be able to detect paraphrased, reused texts, and even semantic synonyms in a definitional sense, thus, demonstrating original thinking and deep comprehension of the material while appreciating the culture.

The contribution of the study can be listed as follows:

1. The paper's first contribution is a dataset of 10,000 Central Kurdish sentence pairs annotated for semantic similarity. The dataset is specifically designed to support plagiarism detection. The data was based on literature, scholarly sources, and web publication.
2. The second contribution is that this data set represents the language and syntax richness of the central Kurdish language, which includes subtle details necessary in the high stakes environments like detecting plagiarism.
3. Third, the recent ones, such as Sentence-BERT (Reimers, 2019) and multilingual BERT (Devlin, 2018) are benchmarked and, as a result, their possibilities as well as their weaknesses are shown. Fourth, an approach is suggested to incorporate this data into the operational plagiarism detectors, closing the gap between theory and practice. Lastly, the dataset will be publicly published, and the world research community will be encouraged to build on it and increase its influence.
4. Fourth, we give a roadmap to the implementation of this dataset in real world application in plagiarism detection systems. Lastly, we make the dataset public and allow the entire research community in the world to utilize it and make it more effective.

The paper is organized in the way that offers a complete overview because, Section 2, touches upon the related work, such as STS, plagiarism detection, and issues that are critical to low-resource language processing. Section 3 guides the reader through the creation of our groundbreaking dataset, detailing the meticulous process of data collection, annotation, and quality checking. Section 4 explains our results and evaluation, including experiment setup, the models and metrics utilized to benchmark this resource. Section 5 explores the implications of the work, Section 6 discusses the limitations and challenges faced during this research, and Section 7 is the exciting direction it leads the way to future research, particularly the creation of Central Kurdish plagiarism detection tools. Finally, Section 8 brings the paper to a conclusion, in which detail our contributions and the revolutionary potential they have for Central Kurdish NLP.

## 2. Related Work

Semantic Textual Similarity (STS) has also been extensively investigated for High-resource languages where there are enough linguistic resources and data available with which robust natural language processing systems can be developed. English is a good example of a high-resource language with sufficient good quality data resources (Center for Democracy and Technology, 2023)However, the progress for low-resource languages has fallen behind since there are few annotated corpora and language-specific resources (Thakur et al., 2020). The status quo of STS research is discussed here with emphasis placed upon its applicability for languages like Central Kurdish, e.g., Persian, Arabic, and Urdu, which are faced with the inherent challenge of low-resource language processing. Some of the challenges are the unavailability of standardized orthography, the scarcity of annotated corpora, and limited computational resources, which complicate the use of natural language processing applications (Ahmadi et al., 2024).

### 2.1 STS in High-Resource Languages

Development of STS datasets such as the SemEval series (Agirre et al., 2015) and resources such as the Microsoft Research Paraphrase Corpus (MRPC) (Dolan and Brockett, 2005) has fueled significant breakthroughs in English and other high-resource languages. Models like Sentence-BERT (Reimers and Gurevych, 2019) and multilingual BERT (Devlin et al., 2019) attain high performance by massive pretraining over diverse corporal. They rely on annotated data and corpora, which are typically not available for low-resource languages.

### 2.2 STS in Persian, Arabic, and Urdu

Persian, Arabic, and Urdu share linguistic and script similarities with Central Kurdish, serving as key reference points for its analysis. These languages utilize scripts derived from the Arabic alphabet and have incorporated numerous Arabic loanwords, leading to shared vocabulary and orthographic features (Hassanpour et al., 2023). These languages exhibit rich morphology, syntactic complexity, and usage of non-Latin scripts, posing similar computational challenges.

1. **Persian:** Persian STS research is relatively recent but has provided resources dataset such as FaSTSim (Abderehman et al., 2022), a Persian sentence pair dataset with semantic similarity annotation. However, Persian remains underrepresented within STS literature when compared with high-resource languages. Transfer learning with multilingual models such as mBERT (Pires, 2019) is encouraging but usually needs fine-tuning with Persian data to capture the language specifics.
2. **Arabic:** Arabic benefits from a richer NLP infrastructure through the aid of resources such as the Arabic Paraphrase Corpus (APC) and the Arabic Semantic Similarity Dataset (Dahy et al., 2021). However, the diversity of the Arabic dialects results in problems akin to those encountered with Central Kurdish, including dialectical differences and script variations. Multilingual model fine-tuning and the creation of dialect-specific resources have proved successful in advancing Arabic STS.
3. **Urdu:** As a low-resource language, limited STS work has been done for Urdu. The Urdu Paraphrase and Semantic Similarity Corpus (Iqbal et al., 2024) is one such focused work within this context. Like Central Kurdish, the Arabic script and complex morphology of Urdu need specialized preprocessing and annotation techniques for obtaining high-quality results.

## 3. Methodology

The methodology is divided into two main parts: corpus creation and model development. Together, these components form the foundation of our approach to introducing STS for Central Kurdish, particularly for plagiarism detection. Figure 1 shows the main architecture proposed method.

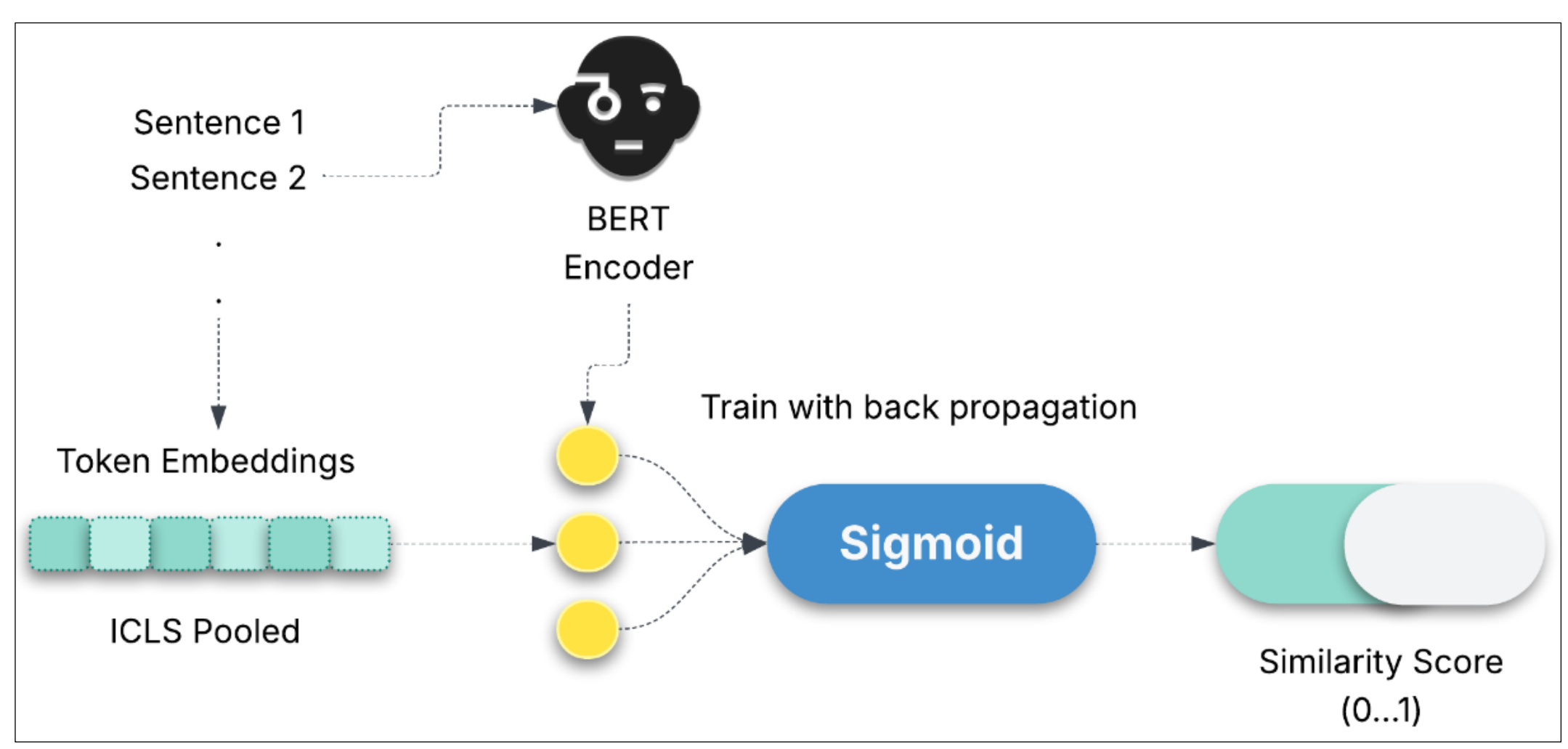

***Figure 1.*** *Main Architecture BERT-Based Sentence Similarity Model: From Token Embeddings to Similarity Score.*

### 3.1 Dataset

This section discusses the datasets developed by PAWS, a high-resource English dataset with high lexical overlap sentence pairs, and the Kurdish Paraphrase Dataset, a low-resource adaptation of PAWS into Central Kurdish, refined for accuracy and cultural relevance. These datasets enhance paraphrase detection across different language resources.

#### *3.1.1 Data PAWS*

PAWS (Zhang et al., 2019) (Paraphrase Adversaries from Word Scrambling) is a dataset designed to challenge and improve paraphrase identification models is primarily in English and is considered a high-resource language (Yang et al., 2019). It includes 108,463 sentence pairs with high lexical overlap but varying paraphrastic relationships, generated using techniques such as controlled word swapping and back-translation. Human annotations ensure grammaticality and accurate labels (paraphrase or non-paraphrase).

The dataset is split into two domains:

1. PAWS-QQP (Quora Question Pairs): 12,665 labeled pairs.
2. PAWS-Wiki (Wikipedia): 65,401 labeled pairs and auxiliary datasets (30k labeled pairs and 656k silver-label pairs).

**Impact:** Training models like BERT with PAWS dataset improve performance from less than 40% to 85% accuracy on challenging paraphrase tasks, demonstrating their value in enhancing model sensitivity to word order and syntactic structure.

***Table 1.*** *Summary of Metadata PAWS Corpus.*

| Category | Details |
| --- | --- |
| Size | 108,463 pairs |
| Generation | Word swapping, back translation |

| Domains | Quora, Wikipedia |
|---|---|
| Key Stats | 12.6k pairs (PAWS-QQP); 65.4k pairs (Wiki) |
| Labels | Paraphrase (Yes) / Non-paraphrase (No) |
| Auxiliary Data | 30k labeled, 656k silver pairs (Wiki) |
| Model Impact | Accuracy improved to 85% (BERT) |

*3.1.2 Kurdish Paraphrase Dataset: Translation and Cultural Adaptation*

To create a Kurdish paraphrase dataset, the PAWS dataset (Zhang et al., 2019) is translated into Central Kurdish using the Google Translate API, followed by a human review to ensure linguistic accuracy and cultural relevance. Human reviewers refine translations to align with Kurdish syntax and semantics, correct machine errors, and adapt cultural references for Kurdish-speaking audiences. Shown summarized in Table 2.

***Table 2.*** *Example English-to-Kurdish Translations with Cultural Adaptations.*

| **Original English Pair** | **Translated and Culturally Adapted Kurdish Pair** |
|---|---|
| Flights from New York to Florida. | فڕۆکەکان لە هەولێر بۆ سلێمانی. |
| Which is the cheapest flight from NYC to Florida? | کە هەرزانترین گەشتەی فڕۆکە لە هەولێر بۆ سلێمانی؟ |
| Can a bad person become good? | ئایا مرۆڤی خراپ دەتوانێت ببێتە مرۆڤێکی باش؟ |
| Thanksgiving dinner was the best event of the year. | ئێوارەخوانی جەژنی نەورۆزی کوردان باشترین بۆنەی ساڵ بوو. |

This process ensures the development of a high-quality dataset tailored to Kurdish linguistic and cultural contexts, improving NLP applications for Kurdish-speaking users.

## 3.2 Central Kurdish Tokenizer

Tokenization is a fundamental step in NLP, especially for morphologically rich languages like Central Kurdish. We developed a Central Kurdish Tokenizer to accurately segment text into tokens. This tokenizer serves two primary purposes: (1) preparing data for training the BERT model and (2) ensuring correct tokenization for input sentences during inference, as shown in Figure 3. To address the "out-of-vocabulary" (OOV) problem, where models struggle with unseen words, we employed WordPiece tokenization has been employed (Schönle et al., 2024), which breaks words into sub word units (Nayak et al., 2020). This approach captures morphological and contextual features, even for previously unseen words. Examples include:

- "نەخێررررر" → Tokens: "نەخ", "##ێر", "##ررر"
- "بەڵێێێێێێێێێێێ" → Tokens: "بەڵ", "##ێێێ", "##ێ"

This technique effectively handles informal writing styles, commonly found in social media texts, ensuring that models capture semantic and emotional nuances.

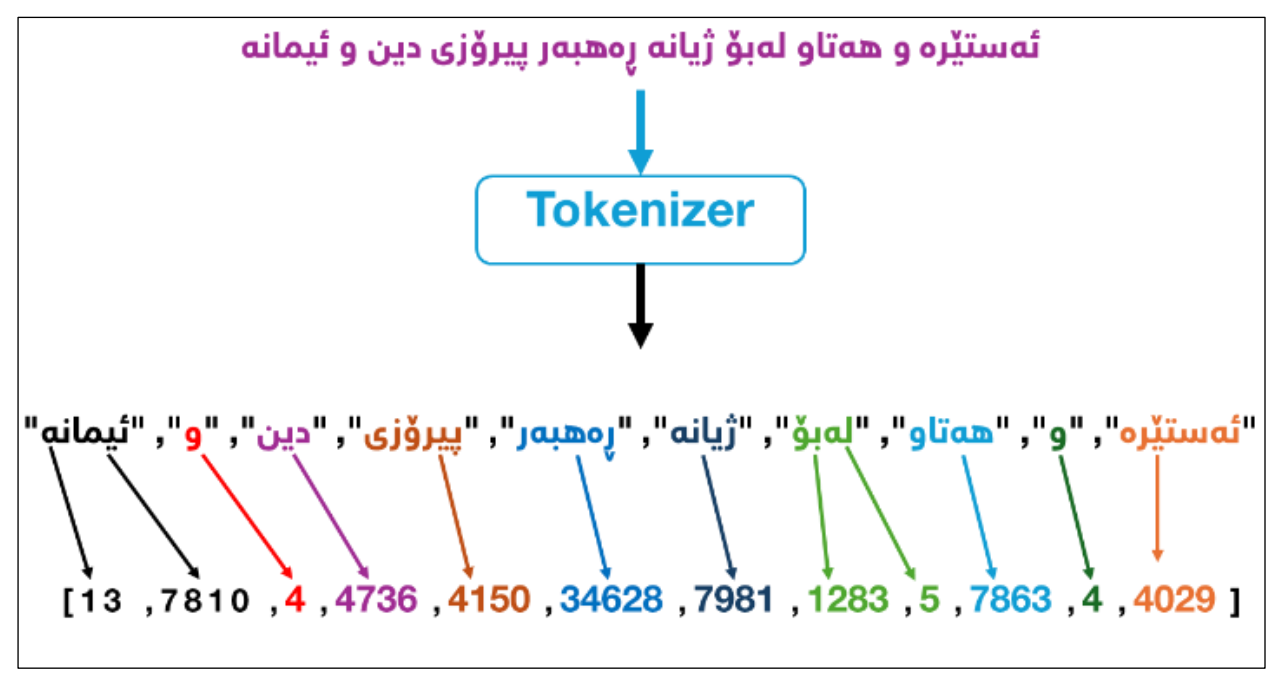

***Figure 2.*** *Example of Kurdish Sentence tokenizer.*

### 3.3 Central Kurdish S-BERT Model

The Central Kurdish S-BERT model adapts the structure of Sentence-BERT (S-BERT) (Reimers, 2019) to fine-tune semantic embeddings specifically for Central Kurdish. Below, we expand on the underlying mechanisms, including word embeddings, transformer architecture, and the attention mechanism. Figure 4 shows the general architecture BERT for STS Task.

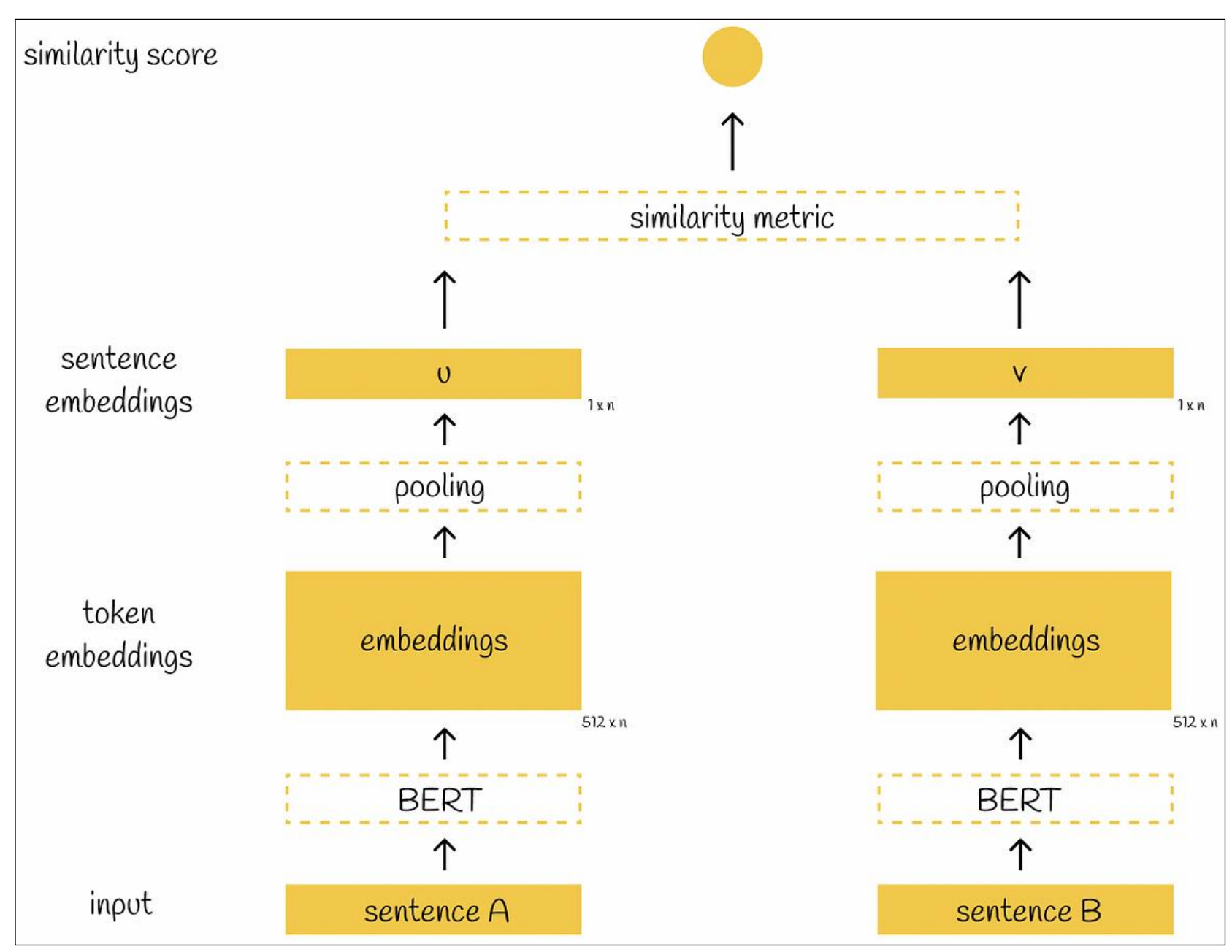

***Figure 3.*** *S-BERT architecture for STS task.*

#### *3.3.1 Architecture Overview*

S-BERT extends BERT (Bidirectional Encoder Representations from Transformers) to generate sentence-level embeddings as shown in Figure 4.

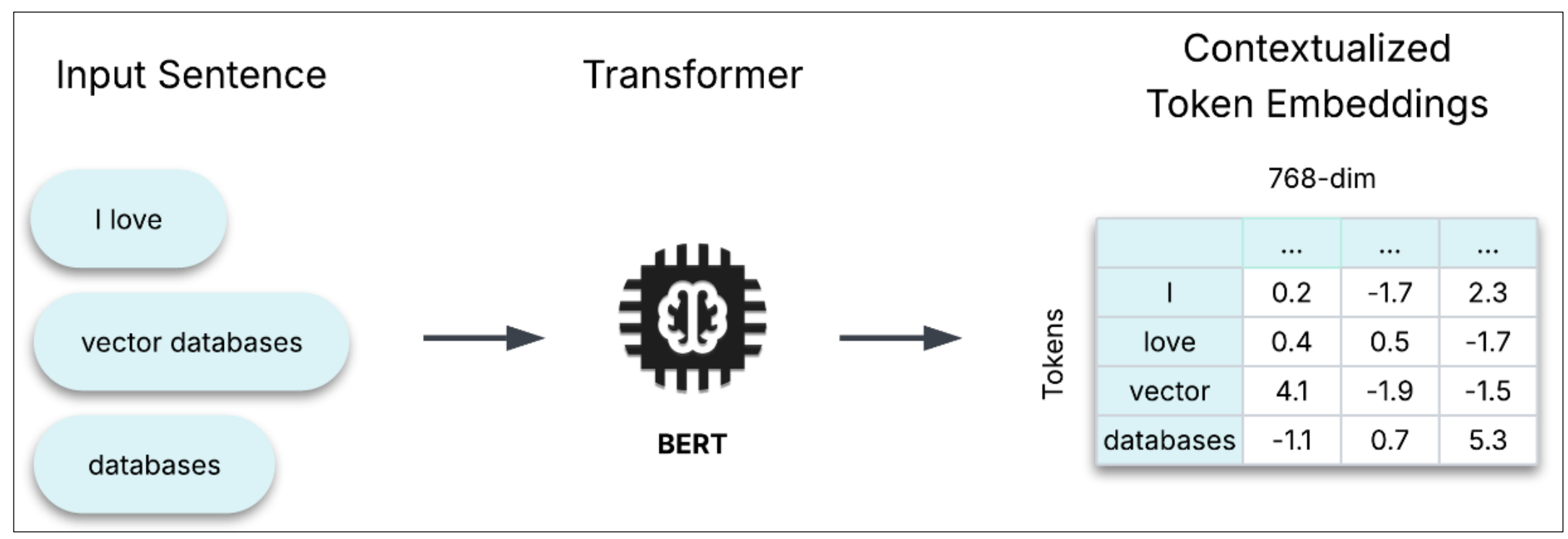

***Figure 4.*** *Contextualized Word Embeddings: Dynamic Word Representations Based on Context.*

Unlike standard BERT, which outputs contextual embeddings for individual tokens, S-BERT combines token embeddings into a fixed-size representation for entire sentences, enabling efficient computation of semantic similarity. The architecture includes the following components, as shown in Figure 5.

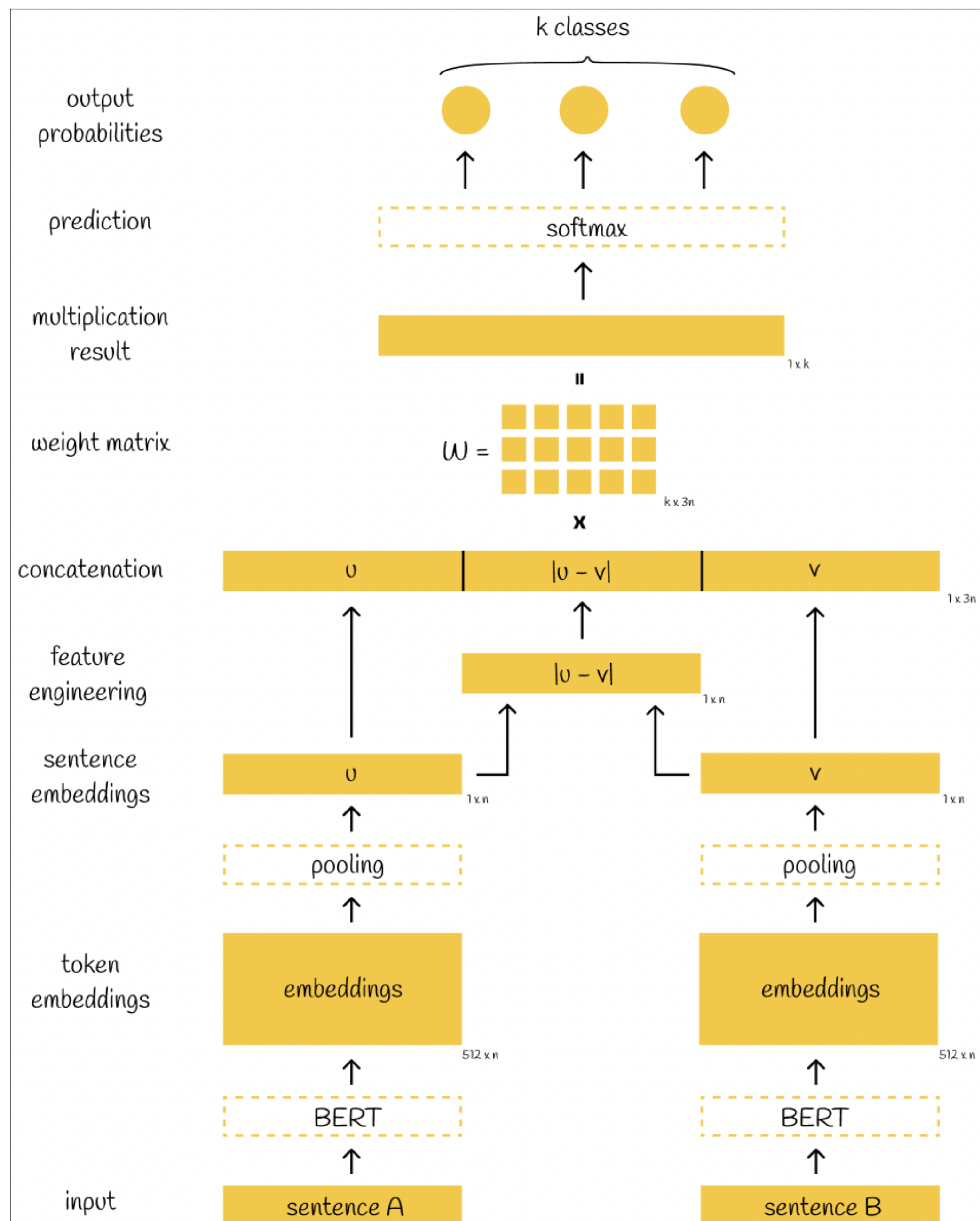

***Figure 5.*** *Sentence Pair Classification Pipeline using BERT Embeddings and Feature Engineering.*

### *3.1.2 Word Embedding*

In transformer-based models, the representation of each input token is constructed by combining multiple embedding vectors that capture different aspects of the token's context. Specifically, the embedding for a token $t$ is computed as the sum of its token embedding, position embedding, and segment embedding, expressed mathematically as:

$$E(t) = E_{token}(t) + E_{position}(t) + E_{segment}(t) \tag{1}$$

Here, $\mathrm{E}_{\mathrm{token}}(t)$ encodes the semantic meaning of the token itself, providing a dense vector that reflects the lexical identity and contextual usage of the word.

Because transformer architectures do not provide a sequential structure, Eposition" (t) the position of the token in the input sequence, is used to provide word order and syntactic detail to the model. Also, Esegment" " (t) differentiates between tokens arising out of diverse segments or sentences and this is especially critical in the context of sentence pairs as in sentence textual similarity (STS) or question answering. The model is built by adding these embeddings, which creates a holistic input representation to the model that combines semantic, positional and segment-level information and thus enables the transformer to productively process and comprehend complex language inputs.

*3.1.3 Self-Attention Mechanism*

The self-attention mechanism is important in the capture of contextual relations of words in a sentence. Given a token t, the input embedding matrix X is linearly mapped to three different representations, namely, Query (Q), Key (K), and Value (V) vectors. These projections are computed as follows:

$$Q = XW_Q, K = XW_K, V = XW_V \tag{2}$$

Where $W_Q, W_K$, and $W_V$ are query, key, and value weight matrices that can be learned respectively. The relevance scores are determined and measure the relevance of each token to others in the sequence; these scores are obtained as the scaled dot product of the queries and keys:

$$Attention(Q, K, V) = Softmax\left(\frac{QK^T}{\sqrt{d_k}}\right)V \tag{3}$$

Here, $QK^T$ represents the dot product of queries and keys and d k is the size of queries and keys to act as a scaling factor to avoid extremely large values which might undermine the training of the model. The softmax function normalizes these scores into attention weights, which are then used to compute a weighted sum over the value vectors $V$. This mechanism enables the model to dynamically attend to and integrate information from relevant tokens across the entire sentence, thereby effectively capturing long-range dependencies and nuanced contextual information.

*3.1.4 Multi-Head Attention*

The multi-head attention mechanism enhances the model's ability to capture diverse contextual relationships by applying multiple attention operations, or "heads," in parallel. Instead of performing a single self-attention operation, the input is linearly projected $h$ times using different learned weight matrices, allowing the model to focus on information from different representation subspaces. For each head $i$, attention is computed using distinct projections of the queries, keys, and values:

$$head_i = Attention\left(QW_{Qi}, KW_{Ki}, VW_{Vi}\right) \tag{4}$$

Where $W_{Qi}, W_{Ki}$, and $W_{Vi}$

do the learned projection matrices depend on the *i-th* attention head. These various attention output streams are then summed and transformed with a final linear transformation with a learnable matrix W O to yield the final output of the multi-head attention layer:

$$MultiHead(Q, K, V) = Concat(head_1, \dots, head_h)W^O \quad (5)$$

Such a method enables the model to collectively consider information across various positions and representation spaces thereby greatly enriching its comprehension of intricate linguistic trends. Multi-head attention combines different views on attention to offer a more powerful and expressive learning mechanism of contextual dependencies on the input sequence.

### *3.1.5 Feed-Forward Network*

After self-attention mechanism, the representation of each token is then refined by running it through a position-wise feed-forward network (FFN). This network works entirely on an individual token and is made up of two linear transformations with a non-linear activation function between them. In theory, the feedforward process when applied to an input vector x, which happens to be the output of the attention layer is given by:

$$FFN(x) = ReLU(xW_1 + b_1)W_2 + b_2 \quad (6)$$

In this formulation, $W_1$ and $W_2$ are learnable weight matrices, and $b_1$ and $b_2$ are corresponding bias terms. The ReLU (Rectified Linear Unit) activation function introduces non-linearity into the transformation, allowing the model to learn complex mappings. This feed-forward network is applied identically and independently to each position in the sequence, enabling the model to process and transform the token representations in a way that enhances their expressiveness for downstream tasks. Together with self-attention, this component contributes to the overall representational power of the transformer architecture.

### *3.1.6 Pooling for Sentence Embedding*

As shown in Figure 7, the outputs from the transformer encoder layers are aggregated to produce a fixed-size sentence embedding denoted as u. One common approach is to compute the mean of the hidden states across all tokens in the sequence. Let $h_i$ represent the final hidden state corresponding to the token $i$, and let $T$ be the total number of tokens in the input sentence. The sentence-level embedding is then calculated as:

$$u = \frac{1}{T}\sum_{i=1}^{T} h_i \quad (7)$$

This mean pooling operation ensures that the resulting embedding u captures information from the entire sentence in a balanced manner. Alternatively, some transformer-based architectures employ the hidden state of a special classification token, typically denoted as [CLS], as the sentence embedding. This token is prepended to the input sequence and is specifically trained to summarize the sentence's content, especially in classification or sentence-level prediction tasks. Both aggregation strategies serve to reduce the variable length sequence of token representations into a single, fixed-dimensional vector that can be used for downstream tasks such as classification, similarity comparison, or retrieval.

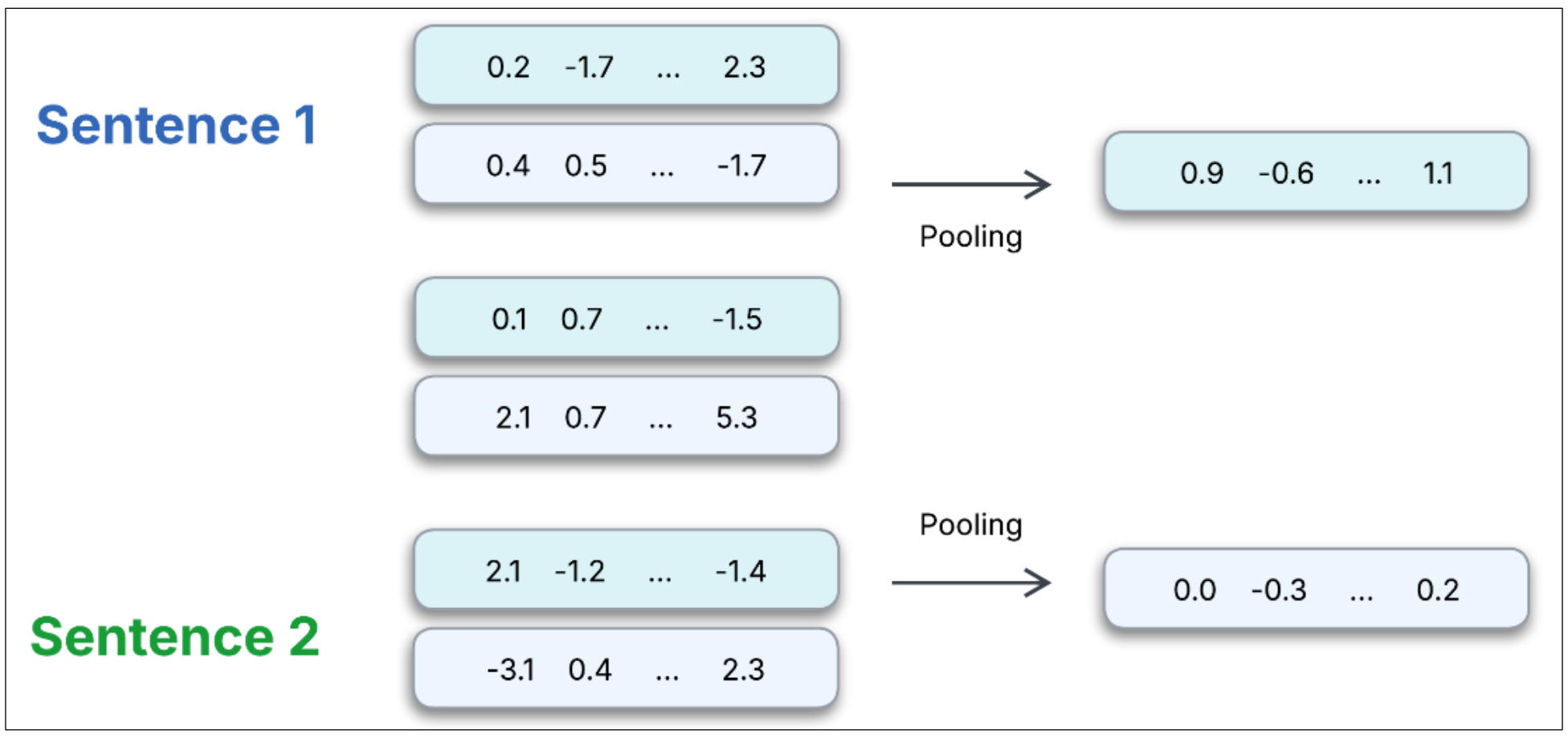

***Figure 6.*** *Sentence Embedding via Pooling of Token Representations.*

### *3.1.7 Cosine Similarity for Sentence Pair Comparison*

As shown in Figure 7, the semantic similarity between two sentence embeddings, u and v, is measured using cosine similarity. This metric quantifies how closely the two vectors align in the embedding space and is defined as:

$$Similarity(u, v) = \frac{u \cdot v}{\|u\|\|v\|} \quad (8)$$

Here, $u \cdot v$ denotes the dot product of the two sentence embeddings, while $\|u\|$ and $\|v\|$ are their respective Euclidean norms. The resulting similarity score ranges from -1, indicating completely opposite meanings, to 1, representing identical semantic content. This similarity measure plays a central role in sentence retrieval and ranking tasks. In the candidate selection step, the goal is to retrieve the top- $k$ most semantically similar sentences from a candidate pool of size $N$. Given a query embedding u and a set of candidate embeddings $\{v_1, v_2, \dots, v_N\}$, the top- $k$ candidates are selected as:

$$C_{top-k} = \mathrm{argmax}_k\{\mathrm{Similarity}(u, v_i) \mid i = 1,2, \dots, N\} \quad (9)$$

This procedure ensures that only the most semantically relevant candidates, based on cosine similarity, are selected for further evaluation or downstream processing.

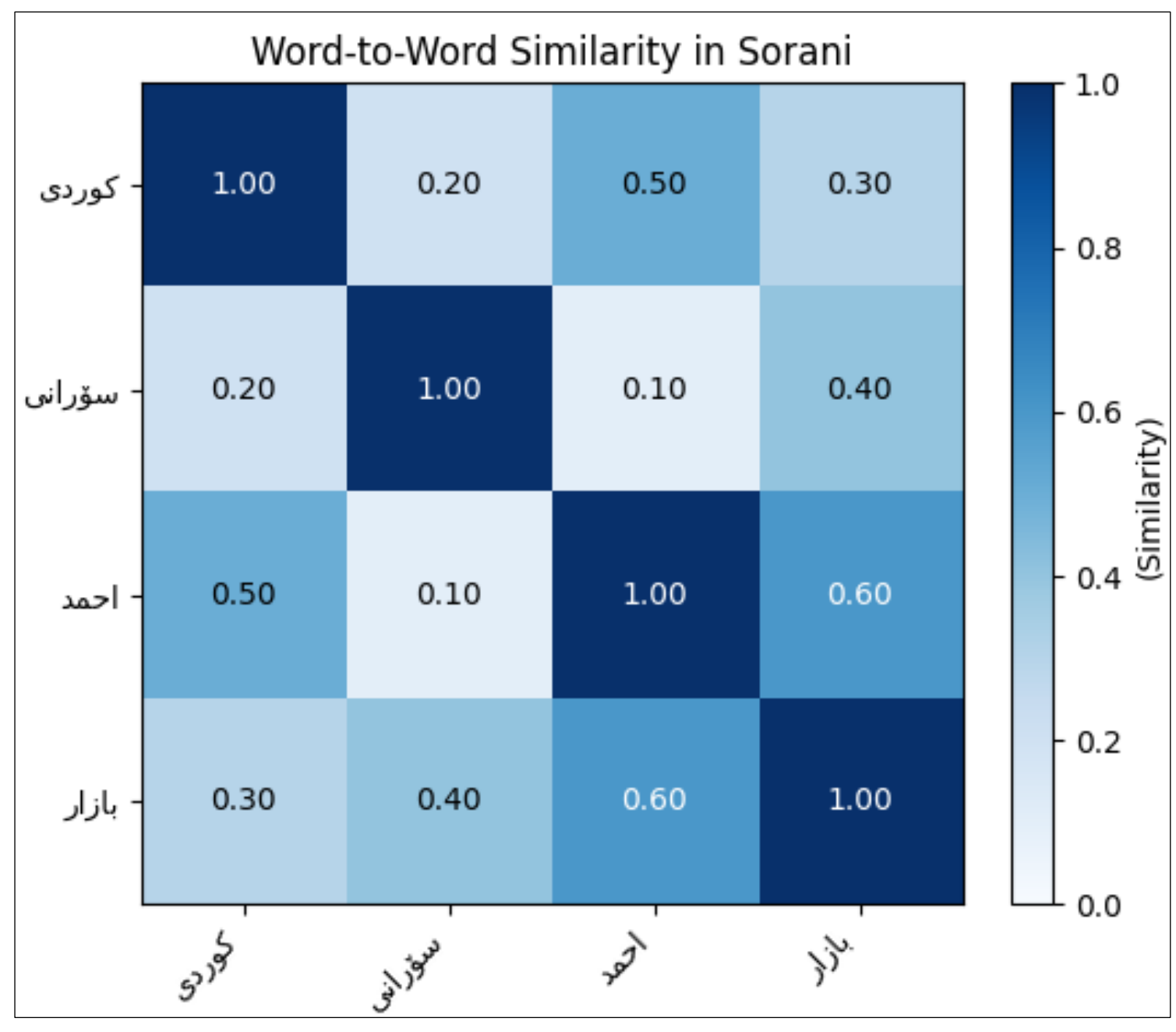

***Figure 7.*** *Word-level Semantic Similarity Matrix: Sentence 1 vs Sentence 2.*

## 4. Evaluation

### 4.1 Semantic Textual Similarity

We thoroughly evaluated the Central Kurdish S-BERT model's performance on the semantic textual similarity (STS) task to analyze its ability to anticipate semantic similarity among various pairs of sentences. This was assessed in both unsupervised and supervised settings (Biggins et al., 2012), with supervised using labeled datasets to learn the extent of similarity and unsupervised using unlabeled data. They benchmarked Central Kurdish S-BERT against S-BERT baseline and multilingual models, indicating the quality of the dataset and the importance of fine tuning for low resource languages.

### 4.2 Experimental Setup

The conducted experiments with the new Central Kurdish Semantic Textual Similarity (STS) dataset that contains 10,000 sentence pairs with annotations of 0-5 for similarity scoring. Extensive care was taken to obtain documents of various types: informal online documents, academic texts, and literature, and this was enough for our purpose. This would ensure a representative sampling of the range of styles, complexity of language structures, and syntactic alternation in Central Kurdish. The evaluation of the three main test models was based on three key criteria. The first was the Rank Correlation Coefficient of Spearman, which analyzes whether there is rank agreement between the human-rated similarity and the model prediction. It ranges from 0 (no correlation) to 1 (perfect rank agreement) and is particularly useful in STS tasks and scenarios where correlation ranking is crucial (Schober, Boer, and Schwarte, 2018).

Second, the average squared differences in the similarity scores predicted and true were measured using Mean Squared Error (MSE), where zero to infinity were the possible values. A smaller value of MSE demonstrates a higher level of performance of the model, whereas the larger values show more significant prediction error (Chai and Draxler, 2014). Finally, Pearson Correlation Coefficient was also reported in completeness. Though it is more sensitive to linear relationships, it gives an extra angle to the alignment of scores, where the values are 0-1 (Zhelezniak et al., 2019). As a measure of similarity, cosine similarity was used as the most basic measurement of sentence embedding comparison. But to be strong, negative Manhattan and Euclidean were also tested.

The outcomes were consistent across all measures, confirming cosine similarity as the most efficient and interpretable metric for this task. A range of baseline models was employed for comparative analysis. Traditional vector space models such as Bag-of-Words (BoW) and Term Frequency-Inverse Document Frequency (TF-IDF), both coupled with cosine similarity, were implemented as foundational baselines. In addition, multilingual transformer models, particularly Multilingual BERT (mBERT) without any fine-tuning on Central Kurdish, were used to assess the zero-shot performance of large pre-trained models. The proposed models included both unsupervised and supervised variants of Central Kurdish Sentence-BERT (S-BERT).

The unsupervised one was trained on a pre-trained S-BERT model, and fine-tuned on no STS-specific data, whereas the supervised one was fine-tuned on the Central Kurdish STS training set. This design allowed comparing the effects of domain-specific fine-tuning on performance in semantic similarity in one of the low-resource language settings directly.

### 4.3 Unsupervised STS Evaluation

The direct use of the Central Kurdish S-BERT and multilingual BERT embeddings without fine-tuning was done for unsupervised evaluation. Cosine similarity was calculated between sentence embeddings and the Spearman correlation with human-annotated scores was computed.

***Table 3.*** *Comparison of Performance Metrics for Kurdish Text Similarity Unsupervised Models.*

| Model | Spearman Correlation | Pearson Correlation | MSE |
|---|---|---|---|
| **Central Kurdish S-BERT** | 0.7 | 0.74 | 0.28 |
| **Multilingual BERT** | 0.67 | 0.65 | 0.32 |
| **TF-IDF + Cosine** | 0.55 | 0.52 | 0.40 |
| **Bag-of-Words + Cosine** | 0.48 | 0.46 | 0.45 |

The Central Kurdish S-BERT achieved the highest performance, significantly outperforming both multilingual BERT and traditional baselines, as shown in Table 3. This finding gives relevance to the tokenization and pretraining approaches depending on the linguistic aspects of Central Kurdish. Multilingual BERT had difficulties with the language morphology and script diversities, and this resulted in reduced correlation scores. Similarly, TF-IDF and Bag-of-Words approaches failed to capture deep semantic relationships, relying only on surface-level word overlaps. In terms of performance analysis, the Central Kurdish S-BERT demonstrated strong capabilities in capturing nuanced semantic similarities, including paraphrases and synonyms, due to its use of contextual embeddings. However, its performance declined on sentence pairs containing idiomatic expressions or heavily informal language, which were underrepresented in the training data.

### 4.4 Supervised STS Evaluation

In the supervised setting, we fine-tuned the Central Kurdish S-BERT on the STS training set using a regression objective function. The fine-tuned model was then evaluated on the test set. Result shown in Table 4.

***Table 4.*** *Comparison of Performance Metrics for Kurdish Text Similarity Supervised Models.*

| Model | Spearman Correlation | Pearson Correlation | Mean Squared Error (MSE) | Spearman Std. Dev. | Pearson Std. Dev. | MSE Std. Dev. |
|---|---|---|---|---|---|---|
| **Central Kurdish S-BERT** | 0.84 | 0.82 | 0.21 | 0.03 | 0.04 | 0.01 |
| **Multilingual BERT** | 0.79 | 0.77 | 0.27 | 0.04 | 0.05 | 0.02 |
| **TF-IDF + Cosine** | 0.62 | 0.59 | 0.44 | 0.05 | 0.06 | 0.03 |
| **Bag-of-Words + Cosine** | 0.55 | 0.53 | 0.48 | 0.06 | 0.07 | 0.04 |

Key observations from the experiments reveal several important findings. Fine-tuning the Central Kurdish S-BERT led to a significant performance improvement, achieving a high Spearman correlation score of 0.84. Compared to multilingual BERT, the fine-tuned Central Kurdish S-BERT outperformed it by 5 percentage points in Spearman correlation, highlighting the effectiveness of dataset-specific fine-tuning. Additionally, the Mean Squared Error (MSE) dropped to 0.21, indicating improved accuracy in similarity prediction. Examples of model predictions further illustrate its strengths. For high similarity pairs, such as "The book is on the table." and "There is a book placed on the table.", the model assigned a predicted similarity score of 4.8, closely matching the gold label of 5. Conversely, for low similarity pairs like "The sky is blue." and "The cat is sleeping.", the model correctly predicted a low similarity score of 0.3, compared to the gold label of 0. These examples demonstrate the model's ability to distinguish both subtle semantic equivalence and unrelated content.

### 4.5 Error Analysis

Although the Central Kurdish S-BERT demonstrated strong overall performance, several challenges persisted. The model struggled with idiomatic expressions unique to Central Kurdish, often producing inaccurate similarity scores. For instance, in the pair "He left with his tail between his legs." and "He walked away quickly.", the predicted similarity was 1.8, while the gold label was 3, indicating a gap in idiom comprehension. Informal texts, particularly those containing social media slang or nonstandard grammar, were also less effectively processed. Furthermore, the model underperformed on domain-specific content such as medical or legal texts, which were underrepresented in the training data.

### 4.6 Comparison with Related Work

When compared to STS evaluations in languages such as Persian, Arabic, and Urdu, the Central Kurdish S-BERT model demonstrated competitive and strong results.

It is remarkable that the Central Kurdish has much fewer linguistic resources and tools available compared to these languages which is one of the reasons why the model has potential despite the lack of support. As an example, the results here were achieved with fine-tuned Persian STS models whose results have been reported to have Spearman correlations of about 0.80. The Arabic STS systems with the use of dialect-specific fine-tuning have reached the levels of correlation of about 0.82, whereas the attempts at the Urdu STS, being limited by the small datasets, reached the correlation of around 0.75. Overall, the Central Kurdish STS dataset with the fine-tuned S-BERT offers a good platform on which to develop natural language processing within the limits of the Central Kurdish language and helps to reduce the gap between it and more well-equipped languages.

## 5. Implications

The high performance of the monitored Central Kurdish S-BERT on STS tasks also shows that it can be applied practically in other areas like plagiarism detection, where checking semantic overlaps in academic and professional texts can take place with the help of automated tools. Also, the embeddings produced by this model provide good opportunities of cross-language usage, such as multilingual search and translation alignment. In the future, the dataset and model can be useful as reliable resources in future studies on low-resource languages to contribute to the topic of unsupervised and domain-specific STS methods. To conclude, it can be stated that the Central Kurdish S-BERT is an effective and strong model that can be utilized to address semantic similarity problems in Central Kurdish. Its good performance is also a new model and creates opportunities to implement NLP in real life, which is a significant step in the language technology in underrepresented linguistic groups.

## 6. Discussion

The findings have significant results and present problem areas of the. applying Semantic Textual Similarity (STS) to Semantic Bert tasks with Central Kurdish S-BERT. Central Kurdish. The outcomes underscore the fact that there is a need to adjust. fine-tuning in low-resource language treatment. Despite outperforming state-of-the-art models and multilingual baselines, the research also identified areas that require future development and research.

### 6.1 Strengths and Key Insights

One outstanding achievement was that the model was able to represent subtle semantic connections between Fine-tuning on the Central Kurdish STS dataset, which enabled it to perform exceptionally in identifying paraphrases, synonyms, and linguistic nuances. This reveals the strength of language-specific datasets in supporting advanced models for underrepresented languages. Whereas multilingual BERT executed well, it faltered with Central Kurdish-specific elements, reaffirming the worthiness of a specially designed S-BERT. The model performed exceptionally well on formal and organized text areas, including academic and literary writing mirroring the diversity and richness of training and test datasets. It shows a robust capacity to understand semantic meaning independently of word overlap distinguishing from traditional techniques such as Bag-of-Words or TF-IDF, demonstrating how transformer-Based embeddings in capturing deep semantic relations. One major takeaway is its real-world applicability and scalability. What is shown by its results is potential for such tasks as plagiarism detection, content alignment, and information retrieval. With high correlation scores, Central Kurdish S-BERT provides a strong platform for building tools that fulfill key academic and career requirements in Kurdish-speaking populations.

### 6.2 Limitations and Challenges

While Central Kurdish S-BERT offers significant advantages, it also faces certain limitations. One major challenge is its difficulty in understanding idiomatic expressions and cultural references. Since many idioms in Central Kurdish depend on context rather than direct linguistic patterns, the model sometimes misinterprets them, leading to lower similarity scores for culturally specific phrases. This highlights the need for a more enriched dataset that includes idiomatic and contextually rich examples. Although Central Kurdish S-BERT has several benefits, it is not without limitations. Perhaps its biggest limitation is its inability to easily understand idiomatic phrases and cultural allusions. As a lot of idioms in Central Kurdish are context-dependent rather than directly linguistic in nature, sometimes they get misunderstood by it, and culturally unique phrases are rated with lower similarity scores. This emphasizes the necessity for a more diverse dataset with idiomatic and contextually rich samples. One limitation is its performance on noisy and casual text, such as that on Twitter and off-the-cuff conversation. The model has difficulty with slang, abbreviations, and sloppy grammar, because of the dataset's bias toward formal writing. To enhance its versatility, future versions will need to include greater samples of casual language.

### 6.3 Comparison with Related Work

After compering with related work for other low-resource languages such as Persian, Arabic, and Urdu, our CKBERT performed well and quite competitive. These languages do however have access to richer linguistic resources and larger data sets, which allows for a wider fine-tuning. Central Kurdish, on the other hand, did not have a large dataset and pre-trained models, which this work set to address. Though these numbers should be considered as two sets of lower and upper bounds respectively, this will at least prove that if enough linguistic resources are made available, Kurdish NLP can be comparable to the state-of-the-art performance in no less resourceful languages in terms of research and practical applications.

### 6.4 Implications for Low-Resource NLP

For low-resource contexts, this achievement has profound implications for NLP. It marks the importance of well-defined datasets, like the Central Kurdish STS dataset which propelled growth despite its dimensions. The study is important for NLP lacking cultural context as evidenced by struggles with idioms and informal expressions, demonstrating a need to weave language and culture into models at word definition levels. Finally, this approach is applicable to other low-resource languages. From dataset creation to model fine-tuning, the steps taken here can be tailored to other dialects with similar challenges, positioning this work as a benchmark in advancing NLP in lesser-known languages.

## 7. Future Directions

Incorporating new samples from both informal and specialized domains will improve the model's performance and accessibility. This includes adding social media posts and domain-specific texts full of metaphors and idioms. In addition, applying multi-task transfer learning with related languages like Persian and Arabic will improve the model's ability to understand nuanced languages. This research can be applied to the development of anti-plagiarism software, semantic search engines, and text summarizers. The Central Kurdish language has not received much technological attention, thus, developing these programs will vastly improve the use of Central Kurdish for scholarly, professional, and online activities. The development of the S-BERT for Central Kurdish represents a significant milestone in the progress of Kurdish NLP technologies. It is clear there are obstacles that must be resolved; nonetheless, this tremendously boosts the prospects of further development. It will allow new services and products to be offered to Kurdish speakers and other users, driven by thoughtful design and deeper understanding of their needs.

## 8. Conclusion

This research achieves an important milestone in language processing for low-resource languages by demonstrating the feasibility of developing high-performing semantic similarity models for neglected languages. We have established a considerable benchmark by STS task evaluation by fine-tuning Sentence-BERT for Central Kurdish using models that target border regions of pragmatic and complex modern NLP tasks, with competitive results that open prospects for practical use and further development in Kurdish NLP. Creating an STS dataset with rich linguistic features is one of the most important contributions of this study. The dataset is composed of a diverse collection of Central Kurdish sentences capturing different dialects and sociolects of the language, containing 10 thousand sentence pairs with annotated semantic similarity. In addition to aiding the research, this dataset has been released to the public, contributing to the development of low-resource language processing.

The performance of the Central Kurdish S-BERT model demonstrates its relevance for practical applications. It significantly surpasses classical methods of semantic equivalence identification like TF-IDF, as well as mBERT, and other multilingual models. Its advanced performance in capturing semantic equivalence makes it useful for Central Kurdish scholarly work, including plagiarism detection, information retrieval, and semantic indexing. Even though this research was very successful, it also uncovered some difficulties. The model was having challenges with informal and conversational phrases such as slang, shorthand, or domain-specific jargon, which illustrates the struggle to keep up with version changes through expansions of alterations made in data collections. Further revisions of the model will require more effort put forth to include diverse forms of speech along with more specialized fields of writing.

These results, in addition to Central Kurdish, serve to advance natural language processing technologies for other languages with lesser-developed resources. This study clearly shows the impact readily available structured datasets and smartly tailored adjustments at the level of underused languages can accomplish. Alongside other language focusses inclusive, this can widen the scope and for NLP systems developed to enable language access within technology, ensuring that every tongue is integrated into the tech world. In the future, we have several possibilities to extend this work. The model adaptability should be improved through a broader dataset containing informal, as well as specialized language. Practical applications such as plagiarism detection and cross-language search engines can also have direct impact on Kurdish speakers. Using transfer learning and multi-task with related languages like Persian or (more distant) Arabic would likely improve the model's performance in overcoming Kurdish-specific linguistic challenges. Overall, this work has gone a long way in providing a sturdy ground for Central Kurdish NLP and has filled a void in computational linguistics. We have enriched semantic similarity modeling to Kurdish by integrating novel methodologies and linguistic and cultural appropriateness from NLP standpoint and proposed a roadmap for future work on low-resource language processing. This work demonstrates

# References

1. Abderehman, m., patidar, j., oza, j., nigam, y., khader, t. A. & karfa, c. 2022. Fastsim: a fast simulation framework for high-level synthesis. Ieee transactions on computer-aided design of integrated circuits and systems, 41, 1371-1385.
2. Abdullah, a. A., veisi, h. & rashid, t. 2024. Breaking walls: pioneering automatic speech recognition for central kurdish: end-to-end transformer paradigm. Arxiv preprint arxiv:2406.02561.
3. Jurafsky, d., & martin, j. H. (2021). Speech and language processing (3rd ed.)
4. Agirre, e., banea, c., cardie, c., cer, d., diab, m., gonzalez-agirre, a., guo, w., lopez-0gazpio, i., maritxalar, m. & mihalcea, r. Semeval-2015 task 2: semantic textual similarity, english, spanish and pilot on interpretability. Proceedings of the 9th international workshop on semantic evaluation (semeval 2015), 2015. 252-263.
5. Azad, n. & rabar fatah, s. 2022. Mulch application and plant spacing influence on growth traits, pests, insects and weeds in cotton (gossypium hirsutum l.) Varieties. Zanco journal of pure and applied sciences, 34, 120-132.
6. Dahy, b., farouk, m. & fathy, k. Assd: arabic semantic similarity dataset. 2021 9th international japan-africa conference on electronics, communications, and computations (jac-ecc), 2021. Ieee, 130-134.
7. Devlin, j. 2018. Bert: pre-training of deep bidirectional transformers for language understanding. Arxiv preprint arxiv:1810.04805.
8. Dolan, b. & brockett, c. Automatically constructing a corpus of sentential paraphrases. Third international workshop on paraphrasing (iwp2005), 2005.
9. El-muwalla, m. & badran, a. 2020. Turnitin: building academic integrity against plagiarism to underpin innovation. Higher education in the arab world: building a culture of innovation and entrepreneurship, 261-268.
10. Iqbal, h. R., maqsood, r., raza, a. A. & hassan, s.-u. 2024. Urdu paraphrase detection: a novel dnn-based implementation using a semi-automatically generated corpus. Natural language engineering, 30, 354-384.
11. Muhamad, s. S., veisi, h., mahmudi, a., abdullah, a. A. & rahimi, f. 2024. Kurdish end-to-end speech synthesis using deep neural networks. Natural language processing journal, 8, 100096.
12. Pires, t. 2019. How multilingual is multilingual bert. Arxiv preprint arxiv:1906.01502.
13. Reimers, n. 2019. Sentence-bert: sentence embeddings using siamese bert-networks. Arxiv preprint arxiv:1908.10084.
14. Schönle, d., reich, c. & abdeslam, d. O. Linguistic-aware wordpiece tokenization: semantic enrichment and oov mitigation. 2024 6th international conference on natural language processing (icnlp), 2024. Ieee, 134-142.
15. Thakur, n., reimers, n., daxenberger, j. & gurevych, i. 2020. Augmented sbert: data augmentation method for improving bi-encoders for pairwise sentence scoring tasks. Arxiv preprint arxiv:2010.08240.
16. Veisi, h., muhealddin awlla, k. & abdullah, a. A. 2024. Kubert: central kurdish bert model and its application for sentiment analysis.
17. Zhang, y., baldridge, j. & he, l. 2019. Paws: paraphrase adversaries from word scrambling. Arxiv preprint arxiv:1904.01130.
18. Khalid, h. S. (2020). Kurdish language, its family and dialects. Kurdiname, 2, 133–146.
19. Center for democracy & technology. (2023). Large language models in non-english content analysis.
20. Ahmadi, s., jaff, d. Q., ibn alam, m. M., & anastasopoulos, a. (2024). Language and speech technology for central kurdish varieties. Arxiv preprint arxiv:2403.01983.

21. Hassanpour, a., sheyholislami, j., & skutnabb-kangas, t. (2023). Kurdish linguistic diversity and multilingualism in the middle east. Journal of applied linguistics, 45(2), 120-135.

22. Schober, p., boer, c., & schwarte, l. A. (2018). Correlation coefficients: appropriate use and interpretation. Anesthesia & analgesia, 126(5), 1763-1768.

23. Chai, t., & draxler, r. R. (2014). Root mean square error (rmse) or mean absolute error (mae)? – arguments against avoiding rmse in the literature. Geoscientific model development, 7(3), 1247-1250.

24. Zhelezniak, v., savkov, a., shen, a., & hammerla, n. (2019). Correlation coefficients and semantic textual similarity. Proceedings of the 2019 conference of the north american chapter of the association for computational linguistics: human language technologies, volume 1 (long and short papers), 951–962.

25. Nayak, n., ang, j., & kuo, c. C. J. (2020). Efficient subword tokenization for neural machine translation and asr. Ieee/acm transactions on audio, speech, and language processing, 28, 530-542.

26. Biggins, s., mohammed, s., oakley, s., stringer, l., stevenson, m., & preiss, j. (2012). University of sheffield: two approaches to semantic text similarity. *first joint conference on lexical and computational semantics (sem), 655–661.

27. Yang, y., zhang, y., tar, c., & baldridge, j. (2019). Paws-x: a cross-lingual adversarial dataset for paraphrase identification. Proceedings of the 2019 conference on empirical methods in natural language processing and the 9th international joint conference on natural language processing (emnlp-ijcnlp), 1298-1308.